\documentclass[runningheads]{llncs}
\usepackage[T1]{fontenc}
\usepackage{graphicx}
\usepackage[table]{xcolor}

\usepackage[separate-uncertainty=true,multi-part-units=single,group-digits=integer,group-minimum-digits=4,detect-all=true]{siunitx}

\usepackage{booktabs}
\usepackage{tabularx}
\newcommand{\maxf}[1]{{\cellcolor[gray]{0.8}} #1}
\newcolumntype{R}[1]{>{\raggedleft\let\newline\\\arraybackslash\hspace{0pt}}m{#1}}

\begin{document}
\title{Do We Really Need Diffusion? A Fast U-Net for Paired Medical Image Translation}

\titlerunning{A Fast U-Net for Paired Medical Image Translation}
%
\author{Alicia Pirwass\inst{1,2,3}\orcidID{0009-0002-1819-3448} \and
Birte Glimm\inst{1}\orcidID{0000-0002-6331-4176} \and
Michael Munz\inst{3}\orcidID{0000-0003-3427-3827} \and Hans-Joachim Wilke\inst{2}\orcidID{0000-0001-6007-8844}}
\authorrunning{A. Pirwass et al.}
%
\institute{Institute of Artificial Intelligence, Ulm University, 89081, Ulm, Germany \and
Institute of Orthopaedic Research and Biomechanics, Centre for Trauma Research, University Hospital Ulm, Helmholtzstrasse 14, 89081, Ulm, Germany \and
AI for Sensor Data Analytics Research Group, Ulm University of Applied Sciences, 89081, Ulm, Germany}
\maketitle              
\begin{abstract}

Magnetic resonance imaging-signal fat fraction (MRI-SFF) quantifies tissue fat and serves as an established biomarker for metabolic and musculoskeletal disorders. The acquisition requires, however, specialized MRI sequences, which are not available routinely. We investigate whether SFF can be estimated from widely available T2-weighted (T2w) MRI via image-to-image translation (I2I). We further compare a lightweight 4-level U-Net to a state-of-the-art Denoising Diffusion Probabilistic Model (DDPM) using a dataset of $\num{230048}$ paired 2D images ($\num{183517}$ train, $\num{23621}$ val, $\num{22910}$ test) from the German National Cohort (NAKO). Both models clearly outperform the identity baseline (Pearson correlation $r = \num{0.769}$, mean absolute error $\mathrm{MAE} = \num{0.070 \pm 0.054}$), which confirms that the models learn a non-trivial cross-modal mapping. Interestingly, the lightweight U-Net outperforms the DDPM in both correlation ($r = \num{0.975}$ vs.\ $\num{0.962}$) and error ($\mathrm{MAE} = \num{0.014 \pm 0.015}$ vs.\ $\num{0.019 \pm 0.019}$), while reducing inference time by a factor of \num{208} (\qty{25.2}{\ms} vs.\ \qty{5227.2}{\ms} per image using 50 Denoising Diffusion Implicit Model (DDIM) steps). The strong clinical performance at substantially reduced computational cost enables real-time clinical use.

\end{abstract}

\section{Introduction}
Magnetic resonance imaging (MRI) can be used to quantify the fat content of tissue, which is an established biomarker for metabolic and musculoskeletal disorders such as sarcopenia (muscle loss e.g.\ due to aging or immobility) and insulin resistance~\cite{li2022sarcopenia}. The signal fat fraction (SFF) measures this fat content as the ratio of the fat MR signal to the total MR signal from fat and water~\cite{reeder2012pdff}. The gold standard for fat quantification is the proton density fat fraction (PDFF), which requires a dedicated multi-echo Dixon MRI sequence: the scanner acquires images at multiple points in time after each excitation pulse, enabling simultaneous correction of physical confounding factors such as magnetic field inhomogeneities and tissue-specific signal decay \cite{reeder2012pdff}. This multi-echo acquisition is technically demanding and time-consuming, and is, therefore, rarely included in routine clinical or epidemiological MRI protocols.

A simpler alternative is the two-point Dixon technique, which acquires only two images at different time points to separate fat and water signal contributions, yielding an uncorrected SFF map. While two-point Dixon is more common than multi-echo Dixon, it is still not consistently acquired across clinical sites and population studies, limiting access to large paired datasets for research.

Standard anatomical MRI sequences such as T2-weighted (T2w) imaging, by contrast, acquire a single image that reflects differences in tissue water content and structure, providing excellent soft-tissue contrast. T2w imaging is among the most frequently acquired MRI sequences worldwide, in clinical practice, research studies, and large epidemiological cohorts, but provides no direct quantitative fat information. Given this abundance, there is considerable research interest in estimating intramuscular fat content from conventional T2w images. However, a recent systematic review~\cite{pirwass2025} shows that existing methods frequently rely on manual or semi-automatic analysis steps, lack validation against quantitative reference standards such as SFF or PDFF, and are therefore difficult to scale or integrate into automated clinical workflows. Image-to-image (I2I) translation offers a promising alternative: by learning a direct mapping from T2w to SFF using paired data, it enables fully automated, quantitative fat estimation from routinely acquired images without requiring Dixon acquisitions at inference time.

We investigate whether this translation is feasible and how complex generative models such as Denoising Diffusion Probabilistic Models (DDPMs) compare to lightweight architectures for this task. We compare a 4-level U-Net to a DDPM on $N = \num{22910}$ paired T2w and Dixon images from the German National Cohort (NAKO) \cite{peters2022,bamberg2015}, a large-scale population-based study with a standardized whole-body MRI protocol that includes a T1-weighted 3D VIBE two-point Dixon sequence covering the neck-to-knee region \cite{bamberg2015}. Since all images were acquired on identical scanners using the same standardized protocol across all study centers, SFF values are internally consistent throughout the dataset. Beyond pixel-wise image quality, we evaluate downstream clinical utility via automated SFF quantification in four paraspinal muscle compartments using VIBESegmentator \cite{graf2025}.

Our main contributions are: (i) the first systematic comparison of U-Net and DDPM for T2w-to-SFF translation on a large-scale epidemiological dataset; (ii) a clinically grounded evaluation via downstream muscle SFF quantification; (iii) we show that, for deterministic paired translation tasks, lightweight models can outperform diffusion models while reducing inference time by a factor of $208$.

\section{Technical Background}
We briefly outline the two architectures compared in this work. For a broader discussion of related approaches see Section~\ref{sec:related}.

\paragraph{U-Net}
The U-Net \cite{ronneberger2015} is a fully convolutional encoder--decoder architecture combining a contracting path for hierarchical feature extraction with a symmetric expanding path for spatial reconstruction. Skip connections between corresponding encoder and decoder levels preserve high-frequency spatial detail, making U-Nets well-suited for paired I2I translation tasks where input and output are spatially aligned.

\paragraph{Denoising Diffusion Probabilistic Models} DDPMs \cite{ho2020} learn to synthesize data by reversing a gradual Gaussian noising process over $T$ timesteps. The model is trained by minimizing a noise prediction objective $\mathcal{L} = \mathrm{E}_{t,x_0,\epsilon} \left[ \| \epsilon - \epsilon_\theta(x_t, t) \|^2 \right]$, where $\mathrm{\epsilon}$ is the added noise and $\epsilon_\theta$ is the learned denoiser. For a full derivation we refer to~\cite{ho2020}. For conditional image synthesis, the input image is provided as additional context at each denoising step, guiding generation towards a target domain. While DDPMs avoid adversarial training instabilities, their iterative sampling procedure results in substantially longer inference times compared to single-pass architectures. For accelerated inference, Denoising Diffusion Implicit Models (DDIMs) \cite{song2021} reformulate the reverse process as a non-Markovian sequence, enabling sampling with substantially fewer steps without retraining the model.

\section{Related Work} \label{sec:related}
In the following, we review related work on quantitative MRI synthesis, I2I translation architectures, and diffusion-based generative models in medical imaging.

\subsection{Quantitative MRI Synthesis}
While prior work has demonstrated the feasibility of inferring PDFF and SFF from alternative MRI sequences, these approaches have focused predominantly on hepatic applications \cite{wang2023deep,anand2024synthesizing,nasir_liver_2025}, smaller cohorts \cite{hess2026rotatorCuff}, or two-point Dixon to PDFF translation \cite{wang2023deep} rather than the T2w-to-SFF setting.

\subsection{Image-to-Image Translation Architectures}
I2I translation refers to the task of learning a mapping between two image domains from paired or unpaired training data. The encoder-decoder architecture introduced by Ronneberger et al.~\cite{ronneberger2015} established the foundation for pixel-wise prediction tasks in medical imaging, combining a contracting path for context aggregation with a symmetric expanding path and skip connections for precise spatial reconstruction. Building on this, Isola et al.~\cite{isola2017} proposed a general-purpose framework for paired I2I translation using conditional adversarial networks, demonstrating that a single architecture with a learned loss function can be applied across diverse translation tasks without task-specific engineering.

In the medical imaging domain, I2I translation has been widely adopted for cross-modal synthesis, including MRI cross-modality translation~\cite{moschetto2025benchmarking,moya-saez_deep_2021,yang_mri_2020}, MRI-to-CT synthesis~\cite{graf_denoising_2023,han_mr-based_2017,lyu_conversion_2022,honey_comparative_2025,hiasa_cross-modality_2018,li_magnetic_2020,li_ct_2023}, CT-to-PET translation~\cite{ben-cohen_cross-modality_2018}, and musculoskeletal applications including muscle segmentation from MRI~\cite{gadermayr_image--image_2021}. These methods typically rely on pixel-wise reconstruction losses (e.g., $\ell_1$) combined with adversarial objectives to encourage perceptual realism \cite{armanious_medgan_2020}. While generative adversarial network (GAN)-based approaches have demonstrated strong performance on paired translation tasks, they can suffer from training instability and mode collapse, which has motivated the exploration of alternative generative architectures.

More recently, Moschetto et al.~\cite{moschetto2025benchmarking} directly compared GAN-based models, DDPMs, and flow matching approaches for T1w-to-T2w MRI translation, finding that the GAN-based Pix2Pix model achieved superior structural fidelity and computational efficiency, suggesting that for deterministic, paired translation tasks, lightweight discriminative models may be preferable to more complex generative frameworks.

\subsection{Denoising Diffusion Probabilistic Models in Medical Imaging}
DDPMs, introduced by Ho et al.~\cite{ho2020}, define a generative process as the reverse of a learned Markov chain that gradually adds Gaussian noise to data. Unlike GANs, DDPMs do not require adversarial training and avoid issues such as mode collapse, which has driven their rapid adoption across generative modeling tasks \cite{kazerouni2023diffusion,luo_review_2025}.

In the medical imaging domain, DDPMs and related score-based models have been applied to a broad range of tasks including image reconstruction, denoising, anomaly detection, and cross-modal synthesis \cite{kazerouni2023diffusion}. For I2I translation specifically, conditional DDPMs have demonstrated strong performance in tasks such as MRI-to-CT synthesis and multi-contrast MRI generation, producing outputs with high anatomical fidelity and favorable convergence properties compared to GAN-based alternatives \cite{khosravi2023ddpm3d,graf_denoising_2023,ozbey_unsupervised_2023}. Despite these promising results, DDPMs carry a fundamental computational disadvantage: inference requires iterating through hundreds to thousands of denoising steps, resulting in substantially longer sampling times compared to single-pass architectures such as U-Nets. Accelerated samplers such as DDIM \cite{song2021} can reduce this to tens of steps while preserving generation quality, though inference remains substantially slower than single-pass architectures. For MRI-to-CT translation, diffusion models have been reported to achieve superior image fidelity despite higher computational cost \cite{honey_comparative_2025}.

An emerging consensus suggests that for deterministic, paired translation tasks, this computational overhead outweighs quality gains \cite{moschetto2025benchmarking}. Rassmann et al.~\cite{rassmann_2026_regression} provide the most systematic evidence to date, proposing YODA, a 2.5D diffusion-based framework that supports regression sampling by directly using the initial single-step denoiser prediction without iterative refinement. They demonstrate that this approach matches or outperforms full diffusion sampling across five diverse datasets including brain MRI and MRI-to-CT translation. They conclude that iterative refinement primarily replicates acquisition noise rather than improving medically relevant information fidelity. Our work extends this line of evidence to the clinically novel T2w-to-SFF translation task on a large-scale epidemiological cohort (\num{22910} test slices), providing a direct comparison between a standard lightweight U-Net and a conditional DDPM without proposing a new framework, and yielding a concrete architectural recommendation for this class of tasks.

\section{Methods}
We next detail our study cohort, image preprocessing pipeline, modeling strategies, and evaluation framework used for SFF estimation from T2w MRI.

\subsection{Dataset and Preprocessing}
The data used in this study were obtained from the German National Cohort (NAKO), comprising $N = \num{30861}$ participants who underwent standardized whole-body MRI. Written informed consent was obtained from all participants and the NAKO study was approved by the responsible ethics committees and conducted in accordance with national regulations and the Declaration of Helsinki (1975, revised version)~\cite{linda_weiser_nako_2025}.

Imaging was performed across all centers using identical scanners (Magnetom Skyra, Siemens Healthcare, Erlangen, Germany) with a field strength of \num{3.0} Tesla and a \qty{70}{\cm} bore, ensuring high consistency across sites.
The imaging protocol included axial T2w HASTE sequences (slice thickness: \qty{5.0}{\mm}; in-plane resolution: $\num{1.4} \times \qty{1.4}{\mm}$; echo time: \qty{81}{\ms}) covering the shoulder-to-epigastric region. In addition, axial T1-weighted 3D VIBE two-point Dixon sequences (slice thickness: \qty{3.0}{\mm}; in-plane resolution: $\num{1.4} \times \qty{1.4}{\mm}$; repetition time: \qty{4.36}{\ms}) were acquired from neck to knee using CAIPIRINHA acceleration. The Dixon technique enables reconstruction of water-only and fat-only images from in- and out-of-phase acquisitions.

From the full cohort, \num{30137} participants with complete T2w HASTE and VIBE Dixon sequences were selected for further analysis. As all sequences were acquired within the same imaging session, no inter-scan registration was required, where registration refers to the process of aligning images from different scans such that corresponding anatomical structures are mapped to the same spatial locations. Spatial alignment between modalities was ensured along the z-axis, defined as the axis orthogonal to the transverse plane. All images were center-cropped from a resolution of $\num{260} \times \num{300}$ to $\num{256} \times \num{256}$ pixels.
SFF maps were computed from the Dixon fat-only and water-only images using a pixel-wise formulation: $\text{SFF} = (\text{fat})/(\text{fat}+\text{water})$. Subsequently, the T2w input images were normalized to the range $[0,1]$ using per-sample min--max normalization. Every second SFF slice was paired with the corresponding T2w HASTE slice to form aligned input–target pairs.

The dataset was split into training, validation and test subsets comprising \num{183517}, \num{23621} and \num{22910} slices, respectively. The split was stratified w.r.t.\ age, body mass index (BMI), and sex to ensure balanced distributions across subsets.

\subsection{Models}
We compare three modeling approaches of increasing complexity for SFF estimation from T2w MRI: a non-learning identity baseline, a lightweight U-Net for deterministic I2I translation, and a DDPM as a representative of modern generative approaches. All models were trained on a single NVIDIA GeForce RTX 4080 GPU (16 GB VRAM).

\medskip
\noindent\textbf{Identity Baseline}
As a simple reference, we consider an identity baseline in which the input T2w HASTE image is directly interpreted as a proxy for SFF. This baseline establishes an upper bound on the achievable error when no cross-modality translation is performed and serves as a lower-complexity reference for evaluating learned models.

\medskip
\noindent\textbf{Image-to-Image U-Net}
We employ a lightweight U-Net with a 4-level encoder–decoder architecture and skip connections. Each convolutional block consists of batch normalization and ReLU activations, and the final layer uses a sigmoid activation to produce outputs in the range $[0,1]$.

The model was trained for \num{30} epochs using the AdamW optimizer with a learning rate of \num{2d-4} and an $\ell_1$ loss. Mixed-precision training (16-bit floating point, FP16) was used with a batch size of \num{16}. The network comprises approximately \num{7.78d6} trainable parameters.

\medskip
\noindent\textbf{DDPM}
The DDPM backbone follows the U-Net architecture proposed by Ho et al.~\cite{ho2020}, with a base channel dimension of \num{64} and channel multipliers $(1, 2, 4, 8)$, resulting in four resolution levels. The conditional T2w input image is concatenated channel-wise to the noisy target at each denoising step. The model predicts both the noise $\epsilon$ and the posterior variance, which is learned via a Kullback--Leibler (KL) divergence-based auxiliary loss weighted by $\lambda = 10^{-3}$~\cite{ho2020}. Training was performed for \num{90} epochs using the Adam optimizer~\cite{ho2020} with a learning rate of \num{2e-5}, $T = \num{1000}$ diffusion timesteps, and a cosine noise schedule with learned variance. Due to memory constraints, a batch size of \num{4} with gradient accumulation over \num{8} steps was used, yielding an effective batch size of \num{32}. Model selection was performed via early stopping on a held-out validation set of \num{800} samples, evaluated every \num{5} epochs using DDIM sampling with \num{50} steps (chosen for computational efficiency during training). Training was stopped after \num{5} consecutive validation epochs without improvement, and the best checkpoint was selected for evaluation. At inference, sampling was performed using the DDIM scheme~\cite{song2021} with \num{50} steps as the standard configuration, which allows accelerated sampling without retraining. Higher step counts were not evaluated, as DDIM sampling has been shown to produce outputs of comparable quality to full DDPM sampling at substantially reduced step counts~\cite{song2021}. The model contains approximately \num{3.263e7} trainable parameters. The larger parameter count of the DDPM reflects its backbone architecture, which is designed for iterative denoising rather than direct regression, and is, therefore, not directly comparable to the U-Net parameter count.

\subsection{Evaluation}
We evaluated the models across four complementary aspects: pixel-wise image quality metrics computed globally and within the body region, downstream SFF quantification in four paraspinal muscle compartments as a clinically relevant task, qualitative visual inspection of representative slices, and inference time as a measure of practical deployability.

\medskip
\noindent\textbf{Image Metrics} Global pixel-wise metrics were computed over all test slices ($N = \num{22910}$), comprising mean absolute error (MAE), structural similarity index measure (SSIM), peak signal-to-noise ratio (PSNR), and pixel-wise bias. Masked metrics were additionally computed within a body mask derived from a fixed intensity threshold of 10\% of the per-image maximum signal intensity on the T2w HASTE images, followed by binary hole filling, combined with VIBESegmentator masks using pixel-wise logical OR. Masked SSIM was computed over the tight bounding box of the body mask to avoid boundary artifacts from zero-padded background pixels. Masked metrics are considered most clinically relevant, as they focus on anatomically meaningful regions.

\medskip
\noindent\textbf{Downstream Task}
We quantified the mean SFF in four paraspinal muscle regions (autochthonous left/right and iliopsoas left/right; labels 59--62) segmented automatically using VIBESegmentator~\cite{graf2025}. Performance was assessed via MAE, bias, and Pearson correlation coefficient ($r$).

\medskip
\noindent\textbf{Qualitative Assessment}
For qualitative assessment, a representative slice was visualized in two views: the full image and a cropped view restricted to the autochthonous and iliopsoas muscle compartments. This progressive masking was chosen to facilitate visual assessment of model differences in the regions relevant to the downstream task.

\medskip
\noindent\textbf{Inference Time Measurement}
Inference times were measured for a single slice with batch size \num{1}, averaged over \num{200} runs after \num{50} warmup iterations with GPU synchronization on an NVIDIA RTX 4080. For the DDPM, inference was performed using the DDIM sampling scheme \cite{song2021}, which allows accelerated sampling by reducing the number of denoising steps without retraining.

\section{Results}
We present the results in the following order: image quality metrics, downstream muscle SFF quantification, qualitative visual inspection, and inference time.

\begin{table}[tb]
\caption{Quantitative image quality metrics (mean $\pm$ std) on the test set
($n = \num{22910}$ slices); global: all pixels; masked: background pixels excluded; best values shaded; all U-Net vs.\ DDPM differences are statistically significant ($p < \num{0.001}$, Wilcoxon signed-rank test)}
\label{tab:image_metrics}
{\setlength{\tabcolsep}{1mm}
\begin{tabularx}{\textwidth}{X X *{4}{r}}
\toprule
\textbf{Model} & \textbf{Eval} & \textbf{MAE $\downarrow$} & \textbf{SSIM $\uparrow$} & \textbf{PSNR $\uparrow$} & \textbf{Bias} \\
\midrule
Identity
  & Global  & \num{0.315 \pm 0.050} & \num{0.259 \pm 0.071} & \num{7.33 \pm 0.77}  & \num{-0.286 \pm 0.056} \\
  (T2w)& Masked  & \num{0.282 \pm 0.062} & \num{0.222 \pm 0.077} & \num{9.30 \pm 1.64}  & \num{-0.221 \pm 0.090} \\
\midrule
U-Net
  & Global  & \maxf{\num{0.136 \pm 0.022}} & \maxf{\num{0.437 \pm 0.078}} & \maxf{\num{12.40 \pm 1.03}} & \num{-0.023 \pm 0.022} \\
  & Masked  & \maxf{\num{0.107 \pm 0.029}} & \maxf{\num{0.490 \pm 0.104}} & \maxf{\num{15.66 \pm 2.33}} & \maxf{\num{-0.002 \pm 0.016}} \\
\midrule
DDPM
  & Global  & \num{0.179 \pm 0.027} & \num{0.366 \pm 0.078} & \num{10.40 \pm 0.89} & \maxf{\num{-0.005 \pm 0.031}} \\
  & Masked  & \num{0.142 \pm 0.038} & \num{0.398 \pm 0.109} & \num{13.45 \pm 2.22} & \num{0.005 \pm 0.024} \\
\bottomrule
\end{tabularx}}
\end{table}

\subsection{Image Metrics}
Table~\ref{tab:image_metrics} reports pixel-wise image quality metrics for all three models. The U-Net significantly outperforms the DDPM across all pixel-wise metrics in both evaluation settings (Wilcoxon signed-rank test, $p < \num{0.001}$ for MAE, SSIM, and PSNR, globally and masked). It achieves a global MAE of \num{0.136 \pm 0.022} and a masked MAE of \num{0.107 \pm 0.029}, compared to \num{0.179 \pm 0.027} and \num{0.142 \pm 0.038} for the DDPM, and \num{0.315 \pm 0.050} and \num{0.282 \pm 0.062} for the identity baseline. A consistent improvement is observed in the masked setting over the global setting across all models, reflecting the removal of trivially predicted background pixels. The identity baseline exhibits a strong negative bias of \num{-0.286 \pm 0.056} globally and \num{-0.221 \pm 0.090} in the masked setting, indicating systematic underestimation of SFF when using the T2w image directly. Both learned models show substantially reduced bias; the U-Net achieves a near-zero masked bias of \num{-0.002 \pm 0.016}, while the DDPM achieves \num{-0.005 \pm 0.031} globally and \num{0.005 \pm 0.024} masked.

\begin{table}[tb]
\caption{Downstream task evaluation: mean SFF estimation accuracy in four
muscle groups segmented using VIBESegmentator;
MAE and bias reported as mean $\pm$ std across all test slices
containing the muscle group; l.: left, r.: right; best values are shaded}
\label{tab:downstream}
{\setlength{\tabcolsep}{1.3mm}
\begin{tabularx}{\textwidth}{lX*{4}{r}}
\toprule
\textbf{Model} & \textbf{Muscle Group} & \textbf{$\text{MAE}\downarrow$} & \textbf{Bias} & \textbf{Pearson $r\uparrow$} & \textbf{$n$} \\
\midrule
Identity 
  & Autochthonous l.   & \num{0.067 \pm 0.054} & \num{-0.058 \pm 0.064} & \num{0.794} & \num{21636} \\
  (T2w)& Autochthonous r.  & \num{0.082 \pm 0.060} & \num{-0.075 \pm 0.069} & \num{0.765} & \num{21617} \\
  & Iliopsoas l.    & \num{0.050 \pm 0.035} & \num{-0.047 \pm 0.039} & \num{0.740} &  \num{7252} \\
  & Iliopsoas r.   & \num{0.062 \pm 0.043} & \num{-0.057 \pm 0.049} & \num{0.671} &  \num{7303} \\
  & \textit{All}      & \textit{\num{0.070 \pm 0.054}} & \textit{\num{-0.063 \pm 0.062}} & \textit{\num{0.769}} & \textit{\num{57808}} \\
\midrule
U-Net
  & Autochthonous l.   & \maxf{\num{0.012 \pm 0.011}} & \num{-0.004 \pm 0.016} & \maxf{\num{0.986}} & \num{21636} \\
  & Autochthonous r.  & \maxf{\num{0.013 \pm 0.011}} & \num{-0.006 \pm 0.016} & \maxf{\num{0.986}} & \num{21617} \\
  & Iliopsoas l.    & \maxf{\num{0.018 \pm 0.024}} & \num{+0.003 \pm 0.030} & \maxf{\num{0.878}} &  \num{7252} \\
  & Iliopsoas r.   & \maxf{\num{0.019 \pm 0.022}} & \num{+0.000 \pm 0.029} & \maxf{\num{0.911}} &  \num{7303} \\
  & \textit{All}      & \maxf{\textit{\num{0.014 \pm 0.015}}} & \textit{\num{-0.003 \pm 0.021}} & \maxf{\textit{\num{0.975}}} & \textit{\num{57808}} \\
\midrule
DDPM
  & Autochthonous l.   & \num{0.016 \pm 0.014} & \num{0.006 \pm 0.021} & \num{0.976} & \num{21636} \\
  & Autochthonous r.  & \num{0.017 \pm 0.015} & \num{0.007 \pm 0.021} & \num{0.977} & \num{21617} \\
  & Iliopsoas l.    & \num{0.024 \pm 0.028} & \num{0.012 \pm 0.035} & \num{0.847} &  \num{7252} \\
  & Iliopsoas r.   & \num{0.024 \pm 0.026} & \num{0.008 \pm 0.035} & \num{0.876} &  \num{7303} \\
  & \textit{All}      & \textit{\num{0.019 \pm 0.019}} & \textit{\num{0.007 \pm 0.025}} & \textit{\num{0.962}} & \textit{\num{57808}} \\
\bottomrule
\end{tabularx}}
\end{table}

\subsection{Mean Muscle SFF}
Table~\ref{tab:downstream} reports mean SFF estimation accuracy in four muscle compartments across $n = \num{57808}$ individual muscle masks. The U-Net achieves the highest Pearson correlation overall ($r = \num{0.975}$) and the lowest MAE (\num{0.014 \pm 0.015}), with particularly strong performance in the autochthonous muscles ($r = \num{0.986}$ for both left and right). The DDPM reaches $r = \num{0.962}$ and $\text{MAE} = \num{0.019 \pm 0.019}$ overall, with autochthonous correlations of $r = \num{0.976}$ and $r = \num{0.977}$ for left and right, respectively. Both models show reduced performance in the iliopsoas compared to the autochthonous muscles. For the left iliopsoas, the U-Net achieves $r = \num{0.878}$ compared to $r = \num{0.847}$ for the DDPM. For the right iliopsoas, we find $r = \num{0.911}$ for the U-Net compared to $r = \num{0.876}$ for the DDPM. Regarding bias, the identity baseline exhibits consistent and substantial negative bias across all compartments (overall $\num{-0.063 \pm 0.062}$), indicating systematic underestimation of SFF. Both models show a small overall bias with the U-Net achieving \num{-0.003 \pm 0.021} and the DDPM \num{0.007 \pm 0.025}. Given the narrow absolute differences between models and the spatially averaged nature of the downstream metric, statistical testing was not performed for the downstream task. The identity baseline yields substantially lower correlations across all muscles (overall $r = \num{0.769}$, $\text{MAE} = \num{0.070 \pm 0.054}$).

\begin{figure}[tb]
    \includegraphics[width=\textwidth]{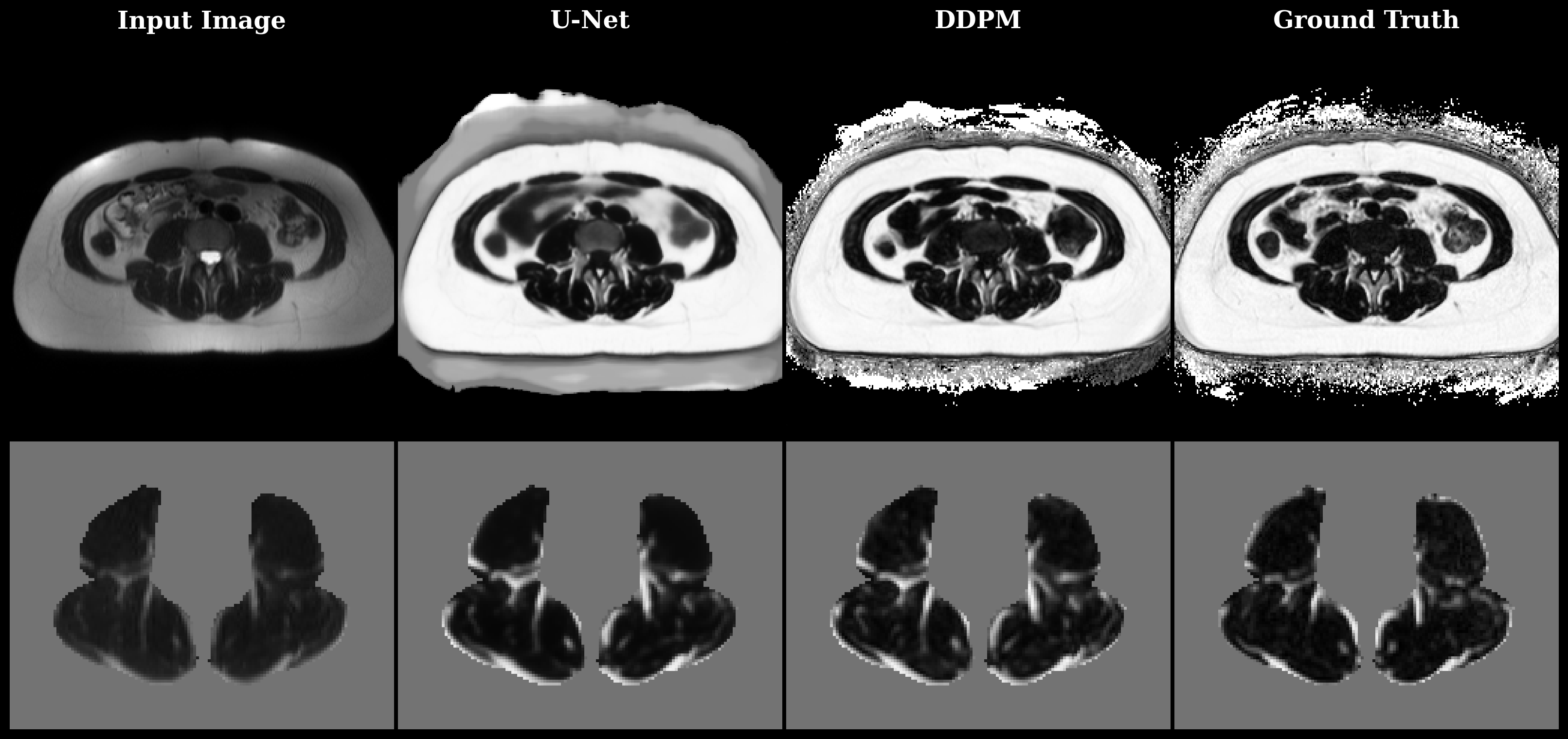}
    \caption{Qualitative comparison of T2w-to-SFF translation. The columns
    show the T2w input image, the U-Net prediction, the DDPM prediction, and
    the SFF ground truth, respectively. The rows display the full
    image and a cropped view of the muscle
    compartments (autochthonous and iliopsoas muscles, left and right)
    segmented using VIBESegmentator; both models reproduce
    the overall fat distribution, while the U-Net prediction appears
    visually comparable to the DDPM at substantially lower inference cost.}
    \label{fig:qualitative}
\end{figure}

\subsection{Qualitative Results}
Figure~\ref{fig:qualitative} shows a representative qualitative comparison of T2w-to-SFF translation for both models alongside the ground truth. In the full image (first row), the DDPM accurately replicates the background noise characteristic of the SFF computation, while the U-Net produces a notably blurred background and visceral region. Focusing on the actual region of interest in the cropped muscle view (second row), both models produce predictions that are visually closely aligned with the ground truth, with no salient differences apparent between the two approaches. A single representative slice is shown; given the test set size of \num{22910} slices, individual examples are inherently selective and the quantitative metrics in Tables~\ref{tab:image_metrics} and~\ref{tab:downstream} provide a more reliable basis for comparison.

\subsection{Inference Time}
Table~\ref{tab:efficiency} reports inference times for both models. The U-Net achieves a speedup factor of \num{208} over the standard DDPM configuration, with inference scaling approximately linearly with DDIM steps.

\begin{table}[tb]
\caption{Model efficiency comparison; inference time measured on an NVIDIA RTX 4080,
batch size \num{1}, averaged over \num{200} runs after \num{50} warmup iterations with GPU synchronization; $^\star$ denotes the standard configuration used in all experiments; best values are shaded, where the U-Net with \qty{7.78}{M} parameters achieves a speedup factor of \num{208} compared to the  standard configuration of the DDPM with its \qty{32.63}{M} parameters}
\label{tab:efficiency}
{\setlength{\tabcolsep}{4mm}
\begin{tabularx}{\textwidth}{X*{3}{r}}
\toprule
\textbf{Model} & \textbf{DDIM Steps} & \textbf{Time/Slice (ms)} & \textbf{50 Slices (s)}\\
\midrule
U-Net & — & \maxf{\num{25.2 \pm 0.1}} & \maxf{\num{1.3}}\\
\midrule
DDPM & \num{10}  & \num{1045.7 \pm 0.6}  &  \num{52.3}\\
DDPM & \num{25}  & \num{2613.8 \pm 1.0}  & \num{130.7}\\
DDPM$^\star$ & \num{50}  & \num{5227.2 \pm 1.4}  & \num{261.4}\\
DDPM & \num{100} & \num{10454.9 \pm 3.2} & \num{522.7}\\
DDPM & \num{200} & \num{20907.8 \pm 2.5} & \num{1045.4}\\
\bottomrule
\end{tabularx}}
\end{table}

\section{Discussion}
This work compares a lightweight U-Net and a conditional DDPM for T2w-to-SFF translation on a large-scale population dataset, evaluating pixel-wise quality and clinically relevant downstream performance. The results consistently favor the U-Net across all evaluated dimensions. We next discuss the implications of these findings with respect to image reconstruction quality, downstream clinical utility, and practical deployability. 

\subsection{Lightweight but Efficient}
The image quality results presented in Table~\ref{tab:image_metrics} demonstrate that the lightweight U-Net consistently outperforms the DDPM across all pixel-wise metrics in both the global and masked evaluation settings, with the exception of bias, where both models achieve near-zero values (\num{-0.023 \pm 0.022} vs.\ \num{-0.005 \pm 0.031} globally; \num{-0.002 \pm 0.016} vs.\ \num{0.005 \pm 0.024} masked). The U-Net achieves a masked SSIM of \num{0.490 \pm 0.104} compared to \num{0.398 \pm 0.109} for the DDPM, and a masked PSNR of \qty{15.66 \pm 2.33}{\dB} versus \qty{13.45\pm 2.22}{\dB}, suggesting that the additional complexity of the diffusion-based sampling process does not translate into improved reconstruction fidelity for this task. While these findings may seem surprising given the U-Net's simplicity, the results can potentially be explained by the perception-distortion tradeoff \cite{blau_perception-distortion_2018}: diffusion models are optimized to sample from the distribution of plausible outputs, which by definition comes at the cost of pixel-wise accuracy. For a paired translation task with a unique ground truth, this distributional objective is misaligned with the reconstruction goal, placing diffusion models at an inherent disadvantage relative to discriminative architectures that directly minimize distortion. Moschetto et al.~\cite{moschetto2025benchmarking} similarly observe that GAN-based and discriminative models can match or outperform diffusion models in deterministic, paired translation settings. Rassmann et al.~\cite{rassmann_2026_regression} confirm this pattern across five diverse datasets, attributing the DM disadvantage to noise replication rather than information loss. The statistical significance of these differences ($p < \num{0.001}$, Wilcoxon signed-rank test) confirms that the U-Net advantage is not attributable to random variation, despite the inherently large test set.

\subsection{Downstream Task}
The downstream evaluation of mean muscle SFF estimation represents the clinically most relevant metric in this work, as it directly reflects the utility of the synthesized SFF maps for quantitative muscle fat assessment. The U-Net achieves an overall Pearson correlation of $r = \num{0.975}$ and MAE of \num{0.014 \pm 0.015}, compared to $r = \num{0.962}$ and MAE of \num{0.019 \pm 0.019} for the DDPM, indicating that both models yield high agreement with the ground truth muscle SFF values. These results compare favorably to previous T2w-to-SFF method comparisons: Pearson correlations of $r = \num{0.87}-\num{0.92}$ for autochthonous muscles were reported for a thresholding-based approach \cite{masi_comparison_2023}, and intraclass correlation coefficient (ICC) values of up to \num{0.88} for the best-performing model (GMM$_\text{3C}$) against a two-point Dixon reference \cite{wesselink_quantifying_2024}. The margin between both models is narrow, suggesting that the performance difference observed in pixel-wise image metrics does not fully translate to the downstream task, likely because mean muscle SFF is a spatially averaged quantity less sensitive to local reconstruction artifacts. The consistently lower performance in the iliopsoas compartments, observed for both models, mirrors the consistently poor psoas agreement reported by Masi et al.~\cite{masi_comparison_2023} and Wesselink et al.~\cite{wesselink_quantifying_2024}, and likely reflects its characteristically low intramuscular fat content and greater anatomical variability resulting in fewer and more heterogeneous training samples. The identity baseline shows substantially degraded downstream performance ($r = \num{0.769}$, MAE $= \num{0.070 \pm 0.054}$) with a systematic negative bias of \num{-0.063 \pm 0.062}, confirming that meaningful T2w-to-SFF translation is required to achieve clinically acceptable muscle SFF estimates and that neither model trivially approximates the ground truth.

\subsection{Clinical Relevance} The U-Net completes a full \num{50}-slice acquisition in \qty{1.3}{\second}, enabling real-time clinical use, compared to \qty{261.4}{\second} for the DDPM at \num{50} DDIM steps (a slowdown factor of \num{208}). Even at the most aggressive reduction to \num{10} DDIM steps, the DDPM still requires \qty{52.3}{\second} per stack, remaining two orders of magnitude slower. Combined with superior pixel-wise image quality and downstream muscle SFF estimation, the computational advantage of the U-Net makes it the more practical choice for deployment in large-scale population studies or clinical settings where throughput and latency are relevant constraints. The DDPM's parameter count of \qty{32.63}{M} compared to \qty{7.78}{M} for the U-Net further compounds this difference in terms of memory footprint, an additional consideration for deployment on clinical hardware.

\subsection{Limitations} This study has several limitations. First, all experiments were conducted on data acquired from a single scanner and protocol within the NAKO cohort, and generalizability to other MRI systems, field strengths, or acquisition parameters remains to be established; notably, DDPMs may offer advantages over discriminative architectures in more heterogeneous multi-site settings where modeling output variability is beneficial. Second, the ground truth used in this work is SFF derived from a two-point Dixon sequence, rather than true PDFF. SFF is susceptible to confounding factors, which are corrected in multi-echo Dixon-based PDFF but not in two-point Dixon acquisitions \cite{reeder2012pdff}. While the use of a standardized single-scanner protocol ensures internal consistency of SFF values across the dataset, future work should investigate whether models trained on SFF generalise to true PDFF targets, and whether the residual confounding in SFF propagates into downstream quantification errors. Third, the two models were trained under different procedures: the U-Net was trained for a fixed \num{30} epochs, while the DDPM was trained with early stopping based on periodic evaluation on a random subset of \num{800} validation samples every \num{5} epochs, representing a trade-off between validation reliability and computational cost. The two training procedures are, therefore, not directly comparable in terms of computational budget and a larger validation subset or more frequent evaluation could yield an improved checkpoint selection for the DDPM. Fourth, while the DDPM inference time could in principle be reduced by decreasing the number of DDIM steps, our results in Table~\ref{tab:efficiency} show that even at \num{10} steps the DDPM remains substantially slower than the U-Net, and the quality implications of aggressive step reduction warrant further investigation. Fifth, no GAN-based baseline (e.g.\ Pix2Pix~\cite{isola2017}) was included; while prior work suggests GANs can match or outperform diffusion models in paired settings \cite{moschetto2025benchmarking,rassmann_2026_regression}, the focus of this work was on comparing lightweight discriminative architectures to diffusion models specifically, and the inclusion of a GAN baseline remains a direction for future work. Sixth, the downstream evaluation is limited to the paraspinal muscle compartments segmented by VIBESegmentator, namely the autochthonous and iliopsoas muscles, as these are explicitly provided by the segmentation tool and are frequently assessed in the musculoskeletal imaging literature \cite{pirwass2025}. Extension to additional muscle groups represents a clinically relevant direction for future work. Finally, latent diffusion models \cite{rombach_high-resolution_2022}, which operate in a compressed latent space and may offer a more favorable trade-off between generative quality and inference speed, were not investigated and could be a promising alternative to the pixel-space DDPM evaluated here.

\section{Conclusion}
We investigated whether SFF can be reliably estimated from T2w MRI using I2I translation, and assessed whether the complexity of diffusion-based generative models is justified for this task. Evaluated on a large held-out test set of \num{22910} paired images from the German National Cohort, a lightweight 4-level U-Net consistently outperformed a conditional DDPM across pixel-wise image quality metrics and downstream muscle SFF quantification, while reducing inference time by a factor of \num{208}. Both models substantially outperformed the identity baseline, confirming that a non-trivial cross-modal mapping is learned. For deterministic, paired medical I2I translation tasks with a unique ground truth, our results indicate that lightweight discriminative architectures provide clinically relevant accuracy at a fraction of the computational cost of diffusion models. We therefore recommend lightweight encoder-decoder architectures as the default choice for such settings, reserving generative approaches for tasks that inherently require stochastic sampling or unpaired training data.

\begin{credits}
\subsubsection{\ackname} The authors acknowledge support of the German Research Foundation (DFG) via project no.\ 558083832, reference no.\ WI 1352/31-1 (A.\ Pirwass, H.-J.\ Wilke) and via GRK 3012, project no.\ 520750254, kemai.uni-ulm.de (A.\ Pirwass, B.\ Glimm) as well as support by the Medical Faculty of Ulm University via Bau\-stein project no. L.SBN.0250 (A.\ Pirwass), and by the Federal State of Baden-Württemberg within the Cooperative Research Training Group ``Data Science and Analytics'' (A.\ Pirwass).

This project was conducted with data (Application No. NAKO-733) from the German National Cohort (NAKO) (www.nako.de). The NAKO is funded by the Federal Ministry of Research, Technology and Space (BMFTR) [project funding reference numbers: 01ER1301A/B/C, 01ER1511D, 01ER1801A/B/C/D and 01ER2301\\A/B/C], Federal States of Germany and the Helmholtz Association, the participating universities and the institutes of the Leibniz Association. We thank all participants of the German National Cohort (NAKO) and the staff of this research initiative. Members and affiliations of the NAKO Investigator Consortium can be accessed via www.nako.de/principal-investigators.

\subsubsection{\discintname}
The authors have no competing interests to declare that are
relevant to the content of this article.
\end{credits}

\clearpage

%
%
%
\bibliographystyle{splncs04}
\bibliography{bibliography}

\end{document}